\documentclass{article}
\PassOptionsToPackage{numbers, compress}{natbib}

\usepackage{preamble}
\usepackage[preprint]{neurips_2022}
\usepackage{geometry} 

\usepackage[utf8]{inputenc} 
\usepackage[T1]{fontenc}    
\usepackage{hyperref}       
\usepackage{url}            
\usepackage{booktabs}       
\usepackage{amsfonts}       
\usepackage{nicefrac}       
\usepackage{microtype}      
\usepackage{xcolor}         

\title{Scalable Collaborative Learning via Representation Sharing}
\author{
  Frédéric Berdoz\\
  MIT, EPFL\\
  \texttt{fberdoz@mit.edu} \And
  Abhishek Singh\\
  MIT\\
  \texttt{abhi24@mit.edu} \And
  Martin Jaggi\\
  EPFL\\
  \texttt{martin.jaggi@epfl.ch} \And
  Ramesh Raskar \\
  MIT\\
  \texttt{raskar@mit.edu} 
}

\begin{document}
\maketitle

\begin{abstract}
Privacy-preserving machine learning has become a key conundrum for multi-party artificial intelligence.
Federated learning (FL) and Split Learning (SL) are two frameworks that enable collaborative learning while keeping the data private (on device).
In FL, each data holder trains a model locally and releases it to a central server for aggregation. In SL, the clients must release individual cut-layer activations (smashed data) to the server and wait for its response (during both inference and back propagation). While relevant in several settings, both of these schemes have a high communication cost, rely on server-level computation algorithms and do not allow for tunable levels of collaboration.
In this work, we present a novel approach for privacy-preserving machine learning, where the clients collaborate via online knowledge distillation using a contrastive loss (contrastive w.r.t. the labels). The goal is to ensure that the participants learn similar features on similar classes without sharing their input data. To do so, each client releases averaged last hidden layer activations of similar labels to a central server that only acts as a relay (i.e., is not involved in the training or aggregation of the models). Then, the clients download these last layer activations (feature representations) of the ensemble of users and distill their knowledge in their personal model using a contrastive objective.
For cross-device applications (i.e., small local datasets and limited computational capacity), this approach increases the utility of the models compared to independent learning and other federated knowledge distillation (FD) schemes, is communication efficient and is scalable with the number of clients. We prove theoretically that our framework is well-posed, and we benchmark its performance against standard FD and FL on various datasets using different model architectures.
\end{abstract}

\section{Introduction}
\label{sec:introduction}
Motivated by concerns such as data privacy, large scale training and others, Machine Learning (ML) research has seen a rise in different types of collaborative ML techniques.
Collaborative ML is typically characterized by an orchestrator algorithm that enables training ML model(s) over data from multiple owners without requiring them to share their sensitive data with untrusted parties.
Some of the well known algorithms include Federated Learning (FL)~\cite{McMahan16}, Split Learning (SL)~\cite{Gupta18} and Swarm Learning~\cite{warnat2021swarm}.
While the majority of the works in collaborative ML rely upon a centralized coordinator, in this work, we design a decentralized learning framework where the server plays a secondary role and could easily be replaced by a peer-to-peer network.
As we show in the rest of the paper, the main benefit of our decentralized approach over centrally coordinated ML is the increased flexibility among clients in controlling the information flow across different aspects such as communication, privacy, computational capacity, data heterogeneity, etc.

Most existing collaborative learning schemes are built upon FL, where the training algorithm and/or the model architecture is usually imposed to the clients to match the computational capacity of the weaker participant (or the server). We refer to this property  as \emph{non-tunable} collaboration, since the degree of collaboration is mostly imposed by the server. While this is less of a problem in cross-silo applications (small number of data owners, large local datasets) since the participants can easily find consensus over the best training parameters, it can constitute a strong barrier for participation in cross-device applications (large number of users with small local datasets and low/heterogeneous computational capacity). Indeed, finding consensus becomes harder in that scenario. To alleviate this, we propose a new framework for \emph{tunable} collaboration, i.e., where each participant has near total control over its data release, its model architecture and how the knowledge of other users should be integrated in its own personalized model.

Our main idea is to share learned feature representations of each class among users and to use these representations cleverly during local training. Since the clients can choose which features are aggregated and shared, our framework enables the clients to assign different privacy levels to different samples. Our decentralized approach also ensures that the overall system remains asynchronous and functions as expected even if all but two clients are offline in the whole system. Finally, our framework makes it convenient to account for model heterogeneity and model personalization, since every user can select a subset of peers based on their goals of generalization and personalization. While some of these advantages have been introduced in recent FL based schemes, our framework allows natural integration of such several ideas.
Our work is different from the body of work done in FL due to its decentralized design. Specifically, our system does not use server-level computation algorithms that are directly involved in the training of the model. Nevertheless, for completeness, we experimentally compare it with FL in Section~\ref{sec:experiments}.

\textbf{Our contribution} can be summarized as follows:
\begin{itemize}
    \item We present a new \emph{tunable} collaborative learning algorithm based on contrastive representation learning and online knowledge distillation.
    \item We prove theoretically that our local objective is well-posed.
    \item We show empirically the advantages of our framework against other similar schemes in terms of utility, communication, and scalability.
\end{itemize}

\section{Related Work}
\label{sec:related_work}

\paragraph{Federated Learning} FL is considered to be the first formal framework for collaborative learning. In their initial paper, \citet{McMahan16} introduce a new algorithm called \texttt{FedAvg}, in which each client performs several optimization steps on their local private dataset before sending the updated model back to the server for aggregation using weight averaging. While this approach alleviates the communication cost of the baseline collaborative optimization algorithm \texttt{FedSGD}, it also decreases the personalization capacity of the global model due to the naive model averaging, especially in heterogeneous environments. Several algorithms have been proposed to address these limitations, in particular  \texttt{FedProx} \cite{Sahu18}, \texttt{FedPer} \cite{Arivazhagan19}, \texttt{FedMa} \cite{Wang20FedMa}, \texttt{FedDist} \cite{Ek21}, \texttt{FedNova} \cite{Wang20FedNova}, \texttt{Scaffold} \cite{Karimireddy19} and \texttt{VRL-SGD} \cite{Liang19}. 
Concerning the server update, \citet{Reddi20} introduce federated versions of existing adaptive optimization algorithms like \texttt{Adagrad}, \texttt{Adam} and \texttt{Yogi}, and \citet{Michieli21} present \texttt{FedProto}, where an attention mechanism is used for clever aggregations. The attention coefficients are computed using \emph{prototypes} (i.e., per class averages of last hidden layer activations, a.k.a. feature representations). While our framework also uses such prototypes, it is conceptually very different as we use them directly in the local objective function (and not in the aggregation).\\ 
Although all these algorithms usually improve the convergence rate, they suffer from the same constraints as \texttt{FedAvg}, i.e., homogeneous model architecture for every client, high communication overhead and \emph{non-tunable} collaboration, all potential barriers for participation.

\paragraph{Fully Decentralized Learning} 
The use of a central server in traditional FL constitutes a single point of failure and can also become a bottleneck when the number of clients grows, as shown by \citet{Lian17}. To alleviate these issues, \citet{Vanhaesebrouck16} formalize a new framework where each client participates in the learning task via a peer-to-peer network using gossip algorithms \cite{Shah09, Dimakis10}. In this configuration, there is no global solution and each client has its own personalized model, which enables both personalization and generalization. On the other hand, it creates other challenges about convergence, practical implementation and privacy \cite{Kairouz19}. Moreover, as in FL, the entire model must be released at every communication round, which can constitute a barrier for participation for the same reasons.

\paragraph{Knowledge Distillation}
\begin{figure}[t]
     \centering
         \begin{subfigure}[b]{1\linewidth}
         \centering
         \includegraphics[width=\linewidth]{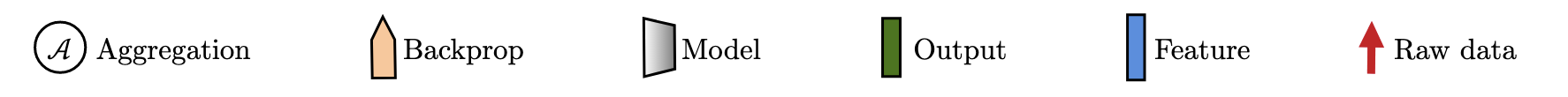}
     \end{subfigure}
     
     \vspace{2ex}
     \begin{subfigure}[b]{0.3\linewidth}
         \centering
         \includegraphics[width=\linewidth]{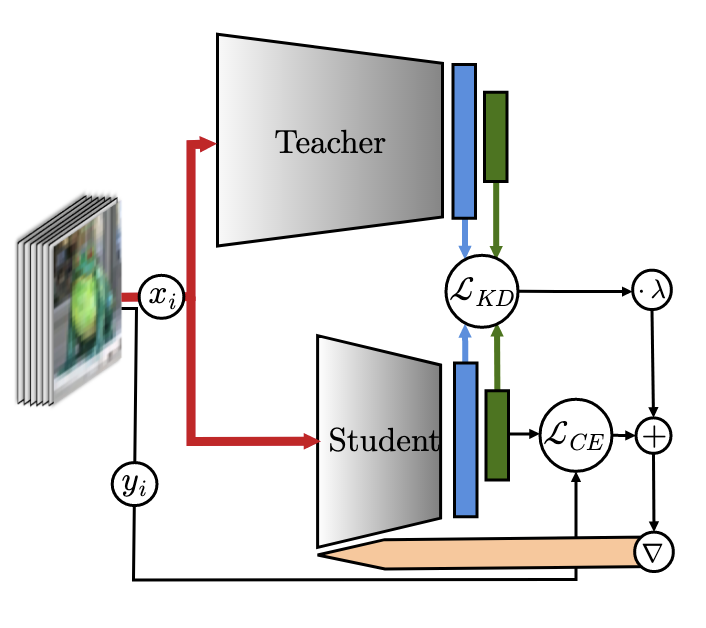}
         \subcaption{Offline KD}
         \label{fig:OffKD}
     \end{subfigure}
     \hfill
     \begin{subfigure}[b]{0.3\linewidth}
         \centering
         \includegraphics[width=\linewidth]{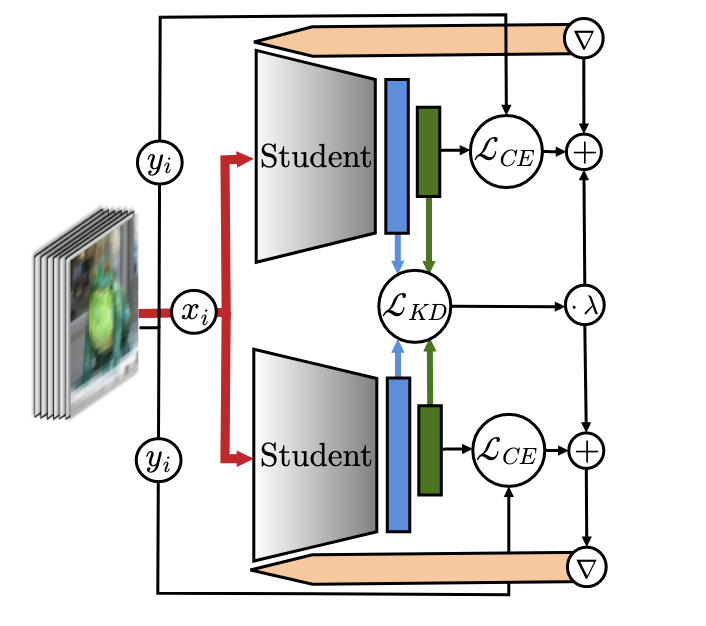}
         \subcaption{Online KD}
         \label{fig:OnKD}
     \end{subfigure}
     \hfill
     \begin{subfigure}[b]{0.3\linewidth}
         \centering
         \includegraphics[width=\linewidth]{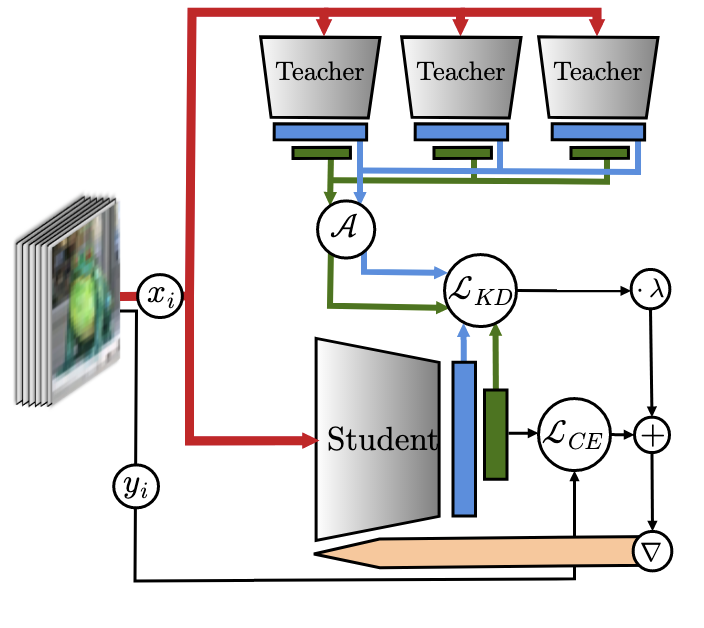}
         \subcaption{Ensemble offline KD}
         \label{fig:OffEKD}
     \end{subfigure}
     \caption{Different types of response-based (green) and feature-based (blue) knowledge distillation.}
\end{figure}
The concept of knowledge distillation (KD) originated in \citet{Bucila06} as a way of compressing models, and was later generalized by \citet{Hinton15} (see \citet{Gou20} for an overview of the field). The standard use case for KD is that of a Teacher-Student (or offline) configuration, in which the teacher model (usually a large and well-trained model) transfers its knowledge to the student model (usually a smaller model) by sharing its last layer activations on a given \emph{transfer} dataset (see Fig. \ref{fig:OffKD}). The knowledge is then \emph{distilled} into the student model using a divergence loss between the teacher and student models outputs (response-based KD) or intermediate layers (feature-based KD) on the transfer dataset. Traditional KD schemes use a transfer set that is similar (or identical) to the teacher training dataset, but some recent work has focused on \emph{data-free} (or \emph{zero-shot}) KD. This can be achieved either by looking at some of the teacher model statistics to generate synthetic transfer data \cite{Lopes17, Nayak19, Bhardwaj19}, or by training a GAN in parallel \cite{Micaelli19, Chen19, Addepalli19}. It has also been shown that positive results can be obtained using mismatched or random unlabeled data for the distillation \cite{Kulkarni17, Nayak20}.
A key concept for our framework is the idea of online KD (or co-distillation \cite{Anil18}, see Fig. \ref{fig:OnKD}), where each model is treated as both a teacher and a student, meaning that the KD is performed synchronously with the training of each model (rather than after the training of the teacher model) \cite{Guo20, Zhang17, Sodhani20}. Finally, ensemble KD (see Fig. \ref{fig:OffEKD}) refers to the setup where the knowledge is distilled from an ensemble of teacher (offline) or teacher-student (online) models.

\paragraph{Collaborative Learning via Knowledge Distillation} A growing body of literature has recently investigated ways of using online knowledge distillation (KD) \cite{Bucila06, Hinton15} for collaborative learning in order to alleviate the need of sharing model updates (FL) or individual smashed data (SL). \citet{He2020} introduce \texttt{FedGKT}, where clients and servers exchange activations in order to consolidate the training. Similarly, \citet{Jeong18} present Federated Knowledge Distillation (FD), where each client uploads its mean (per class) logits to a central server, who aggregates and broadcasts them back. These soft labels are then used as the teacher output for the KD loss during local training. 
A closely related idea is to compute the mean logits on a common public dataset \cite{Li19, Itahara20, Chang19}, but we argue that selecting this dataset can induce bias and is not always feasible, since additional trust is needed for its selection, and sufficient relevant data might be lacking. Besides FD, KD can enable collaborative learning in various ways: In an attempt to decrease the communication cost, \citet{Wu21} introduce \texttt{FedKD}, in which each client trains a (large) teacher network on their private data and transfer locally its knowledge to a smaller student model, which is then used in a standard FL algorithm. On the other hand, \citet{Lin20} and \citet{Chen20} use KD on an unlabeled synthetic or public dataset to make the FL aggregation algorithm more robust. Our approach differs significantly from these schemes, as it does not rely on traditional FL algorithms.
\section{Methods}
\label{sec:methods}
\begin{figure}
    \centering
    \includegraphics[width=1\linewidth]{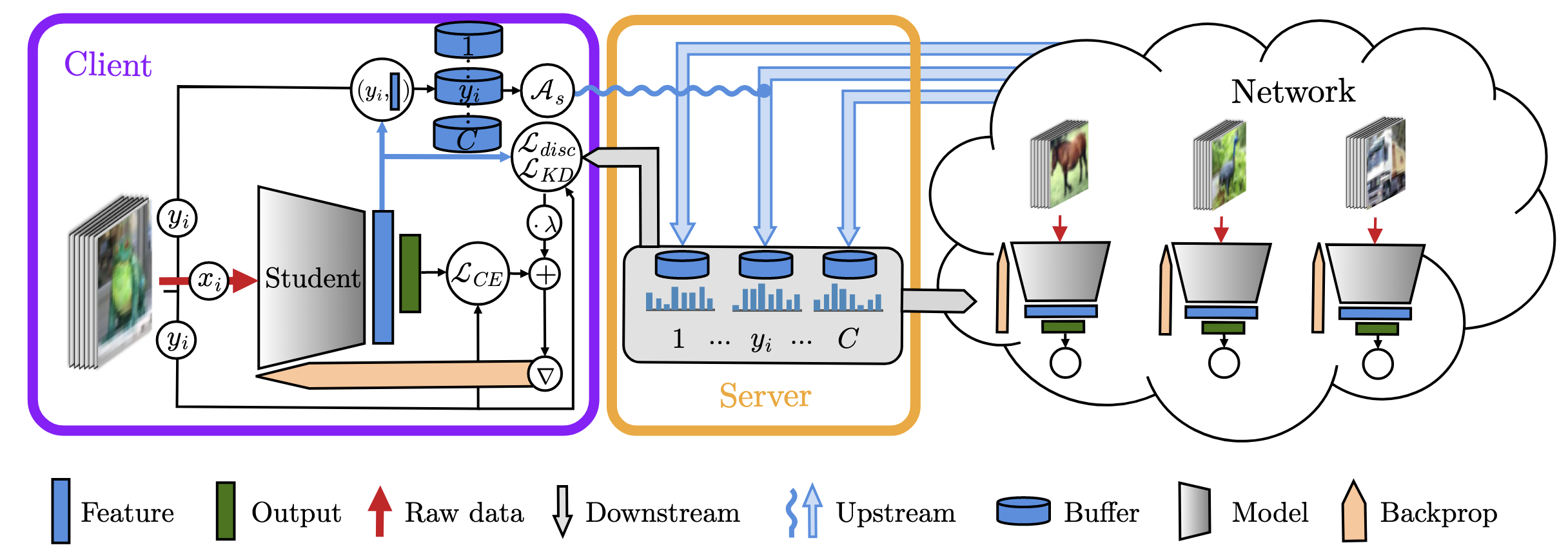}
    \label{fig:FRED}
    \caption{The proposed framework: clients exchange averaged ($\mathcal{A}_s$) feature representations for each class and use these representations in their own local training, but keep their raw data private (on device). The ensemble of participants (network) constitutes the teacher, and each single participant is a student. The server acts mainly as a relay as it does not take part in the training or aggregation of the models. At each communication round, the student downloads a subset of representations and uploads some of its own.}
\end{figure}

\paragraph{Preamble and Motivation} Consider any classification task on $d$-dimensional raw inputs with $C$ distinct classes and a set of $N$ participating users $\{u\}_{u=1}^N$, each with a local private dataset $\mathcal{D}_u$ and a model $f_u$:
\begin{displaymath}
\mathcal{D}_u := \{(\mbf{x}_i, y_i) \overset{iid}{\sim} p_u\}_{i=1}^{n_i}, \qquad p_u(\mbf{x}, y) := p_{\mbf{X}, Y | U=u}(\mbf{x}, y), \qquad f_u = \tau_u \circ \phi_u,
\end{displaymath}
where $p_{\mbf{X}, Y, U}$ represents the joint probability of choosing a user $U$ and drawing a sample $(\mbf{X}, Y)$ from its distribution, and $\tau_u$, $\phi_u$ represent the linear classifier and neural network (up to last hidden layer) of user $u$, respectively (with potentially different architectures across users). More precisely, let $d'$ be the output dimension of $\phi_u$ and $\mbf{w}_u = \{\mbs{\theta}_u, \mbf{W}_u, \mbf{b}_u\}$ be model weights of user $u$. We have:
\begin{align*}
    \phi_u: \R^d & \to \R^{d'} & \tau_u: \R^{d'} & \to \R^C \\
    \mbf{x} & \mapsto \mbf{x}' := \phi_u(\mbf{x}; \mbs{\theta}_u) & \mbf{x}' & \mapsto \mbf{z} := \tau_u(\mbf{x}'; \mbf{W}_u, \mbf{b}_u) = \mbf{W}_u \mbf{x}' + \mbf{b}_u
\end{align*}
where $\mbf{x}, \mbf{x}'$ and $\mbf{z}$ are the raw input, the feature representation and the logits, respectively. We motivate our approach as follows. Assuming no sharing of raw data $\mbf{x}$ (for privacy concerns), any collaborative learning framework falls in one of two buckets: weight sharing or activation sharing. As mentioned in Section \ref{sec:related_work}, sharing weights (i.g., FL) comes with strong constraints (communication, model architecture, etc.) and might not always be suitable. Concerning activation sharing (i.g., SL), it can be done at any layer (hidden or output). However, since activations are usually strongly correlated to the raw input \cite{Vepakomma20}, sharing single sample activations can raise privacy concerns. Instead, sharing averaged activations can easily be met with differential privacy guarantees (see privacy in Section~\ref{sec:methods}). Due to the high non-linearity of neural networks, sharing averaged activations only makes sense at the output layer or last hidden layer (since the classifier $\tau_u$ is linear). Sharing only averaged outputs (averaged over samples of the same class) like in FD has been shown to have limited success as the quantity of shared information is restricted to the dimension of the output $C$, and we argue in this paper that sharing the feature representations (i.e., outputs of the last hidden layer, also averaged over samples of the same class) is more flexible and leads to better results. In this light, our objective is to collaboratively learn the best feature representation for each class, using contrastive (w.r.t. classes) representation learning \cite{Tian19} and feature-based knowledge distillation \cite{Gou20}. In other words, we want to learn collaboratively (i.e., only once) the structure of the feature representation space so that each client does not need to find it on its own with its limited amount of data and/or computational capacity.

\paragraph{Contrastive Objective}
In this section, we introduce a contrastive objective function for private online knowledge distillation (i.e., when users synchronously learn personalized models without sharing their raw input data and by collaborating via online distillation). This new objective function and its derivation are partly inspired by the offline contrastive loss presented in \citet{Tian19} and \citet{Oord18}, with a few important differences. In the offline non-private setting, both the teacher and student models $\phi_t$ and $\phi_s$ have access to the same dataset $\{x_i\}_{i=1}^n$. In that scenario, the representations $\phi_s(x_i)$ and $\phi_t(x_j)$ are \emph{pulled apart} if $i \neq j$ and are \emph{pulled together} if $i = j$. However, in the private setting, $\phi_s$ does not have access to the raw input data that was used to train the teacher model, but has its own private dataset. 
At a given communication round and from the perspective of user $u$, we define $\phi_u$ as the student model and $\phi_U$ as the teacher model, where $U\sim p_U$ is a user selected at random. Consider the following procedure: $u$ samples $Y\sim p_{Y|U=u}$ from its own data distribution  and then samples either jointly (i.e., from the same observation of $Y$) or independently (from two independent observations of $Y$) the two random vectors $\mbs{\Phi}_s$ and $\mbs{\Phi}_t$ defined as follows:
\begin{align}
    \mbf{X} & \sim p_{\mbf{X}|Y,U=u} & \mbs{\Phi}_s &:= \phi_u(\mbf{X}), \\
    U \sim p_{U}, \quad (\mbf{X}_1,..., \mbf{X}_{n_{avg}}) &\overset{iid}{\sim} p_{\mbf{X}|Y,U}, & \mbs{\Phi}_t &:= \frac{1}{n_{avg}} \sum_{i=1}^{n_{avg}}\phi_{U}(\mbf{X}_i).
\end{align}
 The parameter $n_{avg}$ defines over how many samples we take the average, which in turn defines the concentration of the distribution of $\mbs{\Phi}_t$. From the (student) perspective of client $u$ and in the spirit of collaboration, the goal is to maximize the mutual information $\mathcal{I}(\mbs{\Phi}_s, \mbs{\Phi}_t)$. Still from the perspective of $u$, let $p_{s,t}$, $p_s$ and $p_t$ be the joint and marginal distributions of $\mbs{\Phi}_s$ and $\mbs{\Phi}_t$, respectively, and let $I$ be a Bernoulli random variable indicating if a tuple $(\mbs{\Phi}_s, \mbs{\Phi}_t)$ has been drawn from the joint distribution $p_{s,t}$ or from the product of marginals $p_sp_t$. Finally, let $q(\mbf{s}, \mbf{t}, i)$ be the joint distribution of $(\mbs{\Phi}_s, \mbs{\Phi}_t, I)$ such that  $q(\mbf{s}, \mbf{t}|i=1) = p_{s,t}(\mbf{s}, \mbf{t})$ and $q(\mbf{s}, \mbf{t}|i=0) = p_s(\mbf{s})p_t(\mbf{t})$ and suppose that the prior $q(i)$ satisfy $q(i=1) = \frac{1}{K+1}$ and $q(i=0) = \frac{K}{K+1}$, i.e., for each sample from the distribution $p_{s,t}$, we draw $K$ samples from the distribution $p_s p_t$. We can show the following bound.
 \\
\begin{theorem}{} 
\label{thm:I_bound}
Using the above notation, let $h(i, \mbf{s}, \mbf{t})$ be any estimate of $q(i|\mbf{s}, \mbf{t})$ with Bernoulli parameter $\hat{h}(\mbf{s}, \mbf{t})$, and let $\mathcal{L}_{disc}(h, \phi_u)$ be defined as follows:
\begin{equation}
    \mathcal{L}_{disc}(h, \phi_u) :
    = -\E_{(\mbs{\Phi}_s, \mbs{\Phi}_t)\sim p_{s,t}}\left[\log\hat{h}(\mbs{\Phi}_s, \mbs{\Phi}_t)\right] -K\E_{(\mbs{\Phi}_s, \mbs{\Phi}_t)\sim p_sp_t}\left[\log(1-\hat{h}(\mbs{\Phi}_s, \mbs{\Phi}_t))\right].
\end{equation}
The mutual information $\mathcal{I}(\mbs{\Phi}_s, \mbs{\Phi}_t)$ can be bounded as
\begin{equation}
    \mathcal{I}(\mbs{\Phi}_s, \mbs{\Phi}_t) 
    \geq 
    \log(K) - \mathcal{L}_{disc}(h, \phi_u),
    \label{eq:MI_bound}
    \end{equation}
with equality iff $h=q$ (better estimates lead to tighter bounds). The proof is joined in the supplementary material.
\end{theorem}
Hence, by minimizing $\mathcal{L}_{disc}$ in Eq. \eqref{eq:MI_bound}, we optimize a lower bound on the mutual information $\mathcal{I}(\mbs{\Phi}_s, \mbs{\Phi}_t)$.
Taking advantage of the classifier $\tau_u$, a natural choice for $h$ is the discriminator $h_u$ with Bernoulli parameter
\begin{equation}
\hat{h}_u(\mbf{s}, \mbf{t}; \mbf{W}_u, \mbf{b}_u) = \langle \mbox{softmax}(\tau_u(\mbf{s})), \mbox{softmax}(\tau_u(\mbf{t})) \rangle.
\label{eq:contrastive_classifier}
\end{equation}
With this choice, $\hat{h}_u(\mbf{s}, \mbf{t})$ can be interpreted as the estimated probability that the features $\mbf{s}$ and $\mbf{t}$ come from the same class. Note that in their work, \citet{Tian19} train an external discriminator (i.e., that is not defined using the model classifier $\tau_u$).

\paragraph{Final Objective}
Intuitively, from the perspective of $u$, minimizing $\mathcal{L}_{disc}$ ensures that its classifier~$\tau_u$ can distinguish if two feature representations, one local and one from another user, come from the same class. To improve the algorithm convergence and to ensure that $\tau_u$ can classify the feature representation $\mbs{\Phi}_t$ of another client (similar as in Invariant Risk Minimization
 \cite{irm2019}), we also introduce a classical feature-based KD term $\mathcal{L}_{KD}$. This term minimizes the $L_2$ distance between the local and global feature representations of a same class (we define the global representation of class $c$ as the expected value $\E_{(X,U)\sim p_{X,U|Y=c}}[\phi_U(\mbf{X})]$). An important distinction between $\mathcal{L}_{KD}$ and $\mathcal{L}_{disc}$ is that the first one uses an \emph{inter}-client averaged representation, whereas the second one uses an \emph{intra}-client averaged representation.  Combining $\mathcal{L}_{KD}$ and~$\mathcal{L}_{disc}$ with the standard cross-entropy loss $\mathcal{L}_{CE}(\tau_u,\phi_u)$ using the meta parameters $\lambda_{KD}$ and $\lambda_{disc}$, the final optimization problem of $u$  
 becomes:
\begin{equation}
    \mbox{Find}\quad \mbs{\theta}_u^\star, \mbf{W}^\star_u, \mbf{b}^\star_u = \underset{\mbs{\theta}_u, \mbf{W}_u, \mbf{b}_u}{\mbox{argmin}} \, \mathcal{L}_{CE}(\tau_u, \phi_u) + \lambda_{KD}\mathcal{L}_{KD}(\phi_u) + \lambda_{disc}\mathcal{L}_{disc}(\hat{h}_u, \phi_u).
    \label{eq:final_objective}
\end{equation}
\paragraph{Implementation}
With the standard assumption that the datasets are composed of IID samples drawn from their corresponding distributions, the expected losses $\mathcal{L}_{CE}, \mathcal{L}_{KD}$ and $\mathcal{L}_{disc}$ can be approximated by their unbiased mini-batch estimators $L_{CE}, L_{KD}$ and $L_{disc}$, respectively. For one given sample, the loss functions are given by $\ell_{CE}(\mbf{z}, c) := - \log\left(\mbox{softmax}(\mbf{z})\right)_c$, $\ell_{KD}(\mbf{x}', \mbf{x}'') := \Vert \mbf{x}' - \mbf{x}'' \Vert^2$ and finally
\begin{align}
\ell_{disc}: \{0,1\} \times \R^{d'} \times \R^{d'}& \to \R_+ \nonumber \\ 
(i, \mbf{s}, \mbf{t}) & \mapsto - i\log\left(\hat{h}(\mbf{s}, \mbf{t})\right)-(1-i)\log\left(1-\hat{h}(\mbf{s}, \mbf{t})\right).
\end{align}
Finally, for the sampling procedure of the contrastive objective, we use the most intuitive scheme: for every training sample $(\mbf{x}_i, y_i)$ with feature representation $\mbf{s}_i = \phi_u(\mbf{x}_i)$, we use one observation of $\mbs{\Phi}_t$ sampled using $y_i$ (i.e., $I=1$), and one observation of $\mbs{\Phi}_t$ sampled using each $c\neq y_i$ (i.e., $I=0$). Thus, we have $K=C-1$.
With these considerations, a detailed description of our collaborative learning framework can be found in the supplementary material.
\paragraph{Communication}
In terms of communication, the uplink and downlink volumes are of order ${\mathcal{O}((M_\uparrow +1)Cd')}$ and $\mathcal{O}(N(M_\downarrow+ 1)Cd')$ per round, where $M_\uparrow$ and $M_\downarrow$ represent the number of $\mbs{\Phi}_t$ realizations (per class) that are uploaded and downloaded by the clients, respectively (these parameters can be tuned to match the communication capacity of the network). For comparison, these volumes are of order $\mathcal{O}(D)$ and $\mathcal{O}(ND)$ for FL (with $D$ the model size) and $\mathcal{O}(nd')$ and $\mathcal{O}(Nnd')$ for SL. Because in most scenarios $D \gg n \gg d' \gg C$ and since $M_\uparrow$ and $M_\downarrow$ are tunable, we observe that our framework is communication efficient compared to FL and SL.
\paragraph{Relaxation to peer-to-peer}
We emphasize here that our contrastive objective $\mathcal{L}_{disc}$ is fully compatible with a peer-to-peer configuration, since the server only acts as a relay for the observations of $\mbs{\Phi}_t$. For the traditional feature-based KD objective $\mathcal{L}_{KD}$, we use one global representation per class (i.e., one for the whole network), which in theory can only be computed by a central entity. One way to alleviate this could be to use the average of all the observations of $\mbs{\Phi}_t$ that were downloaded by user $u$ as a proxy for the global feature representations. However, to focus solely on the effectiveness of the proposed objectives rather than the topology of the network, we only present experiments in which a central entity computes the global representations.

\paragraph{Privacy}
Our framework naturally attains a certain degree of privacy by sharing representations from the penultimate layer, which, as per information bottleneck principle~\cite{tishby2015deep}, has minimal mutual information from the original data. Although such a level of privacy may not be enough from a membership attack point of view, we emphasize that the privacy is amplified by averaging out these representations, hence amplifying the privacy of individual samples. To further make our framework formally private, we propose using recent progress in differentially private mean estimation~\cite{liu2021robust,brown2021covariance} to obtain a differentially private mean before it gets shared with other clients. Furthermore, our framework naturally dovetails with recent works~\cite{kotsogiannis2020one,acharya2020context} in variable privacy levels for different samples. The users can assign variable levels of privacy to different samples before they get aggregated in a privacy preserving manner.

\newpage
\section{Experiments}
\begin{wrapfigure}{r}{0.4\textwidth}
    \centering
    \includegraphics[width=0.95\linewidth]{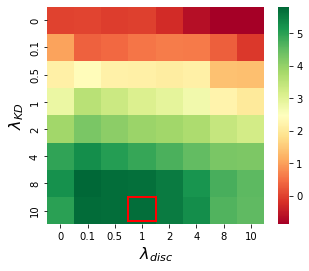}
    \caption{Ablation study for $\lambda_{KD}$ and $\lambda_{disc}$. Average test accuracy improvement [$\%$] w.r.t. IL (upper left corner) when different combinations of $\lambda_{KD}$ and $\lambda_{disc}$ are used (MNIST/LeNet5 experiment with 5 users and 100 communication rounds). The red square indicates the value of $\lambda_{KD}$ and $\lambda_{disc}$ used in all our experiments.}
    \label{fig:lambda_ablation}
\end{wrapfigure}
\label{sec:experiments}
\paragraph{Datasets, models and training} We run several experiments with the MNIST \cite{mnist}, Fashion-MNIST \cite{fmnist} and CIFAR10 \cite{cifar10} datasets. For MNIST, we use a simple convolutional neural network (CNN) similar to LeNet5 ($\approx 30K$ parameters) \cite{lenet5} and for Fashion-MNIST and CIFAR10, we use ResNet9 ($\approx 2,4M$ parameters) and ResNet18 ($\approx 11.3M$ parameters) architectures \cite{resnet}, respectively. For the dimension of feature representation space, we set $d'=84$ for LeNet5, $d'=128$ for ResNet9 and $d'=256$ for ResNet18. In order to simulate a scenario where the data is sparse, we only select a fraction of the train dataset (1200 samples for MNIST, 6000 for Fashion-MNIST and 10000 for CIFAR10) that we split uniformly at random across $N\in\{2,5,10\}$ users. For the validation, we use the entire test dataset for each task (10000 samples). In order to have fair comparisons, we train all the models for the same number of communication rounds, and we stop the training as soon as framework has reasonably converged ($r=100$ for MNIST/LeNet5, $r=20$ for Fashion-MNIST/ResNet9 and CIFAR10/ResNet18). We compare our framework with centralized learning (or CL, i.e., with $N=1$ and $\lambda_{KD} = \lambda_{disc} = 0$), independent learning (or IL, i.e., with $\lambda_{KD} = \lambda_{disc} = 0$), federated learning using \texttt{FedAvg} (FL) and federated knowledge distillation (FD). We use the default learning rate $\eta = 10^{-3}$ for all experiments and we perform 1 local epoch of training per communication round. For CL, IL, FD and our framework, we use the \texttt{Adam} stochastic optimization algorithm \cite{adam}. Finally, supported by Fig. \ref{fig:lambda_ablation}, we set $\lambda_{KD}=10$ and $\lambda_{disc}=1$ in our final objective (Eq.  \eqref{eq:final_objective}).

\paragraph{Network emulation} For our contrastive objective $\mathcal{L}_{disc}$, we use $n_{avg}=10$ (i.e., for every class, each user selects $n_{avg}$ samples of that class, computes and averages their feature representations, uploads them to the server and downloads the representations of another user chosen at random). For the feature-based KD objective $\mathcal{L}_{KD}$, each client average the feature representation of all the samples of a same class and uploads these averaged representations to the server, which in turn averages them to obtain one global representation per class. For simplicity, we use $M_\uparrow =M_\downarrow = 1$ (clients upload and download one observation of $\mbs{\Phi}_t$ for each class).
\paragraph{Software and hardware}
We use the CUDA enabled PyTorch library \cite{pytorch} with a GTX 1080 Ti (11GB) GPU and an Intel(R) Xeon(R) CPU (2.10GHz).

\section{Discussion}
\label{sec:discussion}
\begin{figure}
     \centering
     \begin{subfigure}[b]{0.245\linewidth}
         \centering
         \includegraphics[width=\linewidth]{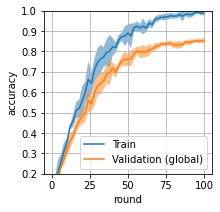}
         \subcaption{IL}
         \label{fig:MNIST_il}
     \end{subfigure}
     \hfill
     \begin{subfigure}[b]{0.245\linewidth}
         \centering
         \includegraphics[width=\linewidth]{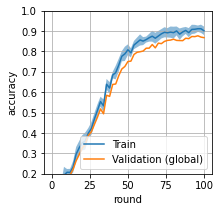}
         \subcaption{FL}
         \label{fig:MNIST_fl}
     \end{subfigure}
     \hfill
     \begin{subfigure}[b]{0.245\linewidth}
         \centering
         \includegraphics[width=\linewidth]{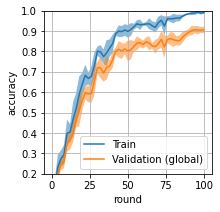}
         \subcaption{FD}
         \label{fig:MNIST_fd}
     \end{subfigure}
     \hfill
     \begin{subfigure}[b]{0.245\linewidth}
         \centering
         \includegraphics[width=\linewidth]{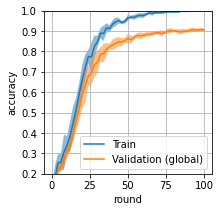}
         \subcaption{Ours}
         \label{fig:MNIST_cfkd}
     \end{subfigure}
     \caption{Comparison of the train and test accuracy during 100 communication rounds of IL, FL, FD and our framework on the MNIST dataset with LeNet5 architectures and $N=5$ users. The shaded areas represent $\pm$ the standard deviation of the metric across clients. For the validation, each model is tested using the entire test dataset.}
     \label{fig:MNIST}
\end{figure}

\begin{figure}
     \centering
    \includegraphics[width=\linewidth]{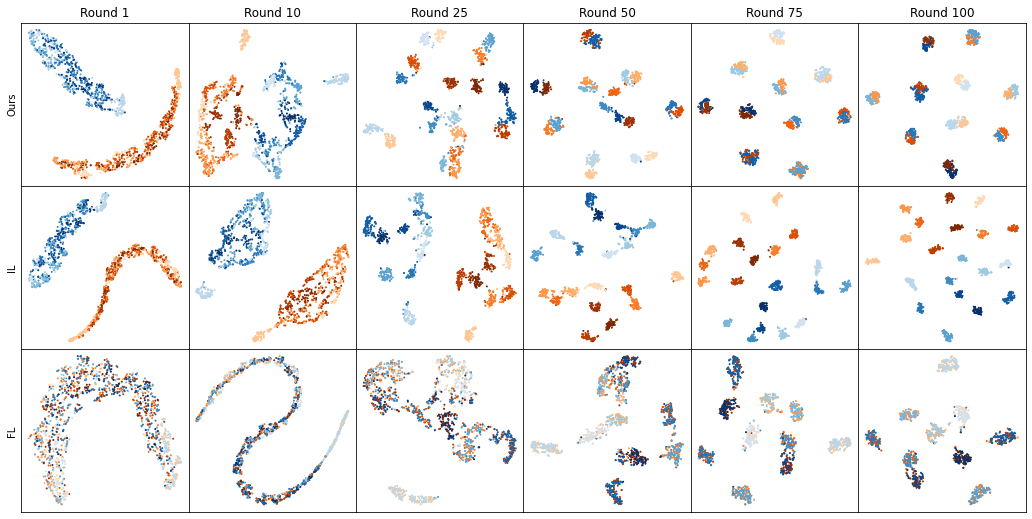}
    \caption{T-distributed Stochastic Neighbor Embedding \cite{tSNE} of the feature representation at different rounds in a two party configuration for the MNIST/LeNet5 experiment. Clients are distinguished by color and classes by opacity.}
    \label{fig:tSNE}
\end{figure}

\paragraph{Utility}
As seen in Table \ref{tab:results}, our framework outperforms every other framework for when a small model is used (MNIST/LeNet5), especially when the number of clients grows, which is typically the kind of configuration that would be relevant for a cross-device application). In that setup, FL performs particularly poorly as it struggles to find a low-capacity model that matches the data distribution of each client. This is particularly visible on Fig. \ref{fig:MNIST_fl}. Although in that configuration, FD shows similar performance for small number of clients ($N=2,5$), it still exhibits a lower rate of convergence (compare Fig. \ref{fig:MNIST_fd} and Fig. \ref{fig:MNIST_cfkd}). Our framework even outperforms centralized learning (CL) when $N=2$, suggesting that the added objectives can also be seen as regularizers. 
For the Fashion-MNIST dataset, our framework shows significant improvement over IL and FD (i.e., frameworks where there are no global model), but is not able to compete with FL anymore, which in that case even outperforms CL. However, the comparison between FL and our framework is unfair for large models due to the amounts of shared information (i.g., for ResNet9, one communication round of our framework communicates $\approx 1000$ times fewer bits than standard FL. This ratio increases to $\approx 2000$ for ResNet19).
Similar conclusions can be drawn for the CIFAR10 experiments. However, in that case, even IL outperforms our framework for $N=10$ after $r=20$ rounds. Indeed, since our objective function is highly complex (the global minimum depends on the data of other peers), the algorithm struggles to converge to the optimal model when the number of parameters is very large. However, we argue that this is not critical since training ResNet18 models on low capacity devices is not particularly adapted to cross-device applications, as they usually require powerful GPUs and large datasets to converge.
\begin{table}[h]
  \caption{Average test accuracy over clients $[\%]$ of the different frameworks after $r$ communication rounds when the same amount of data is divided uniformly at random between $N$ users. We use 1200 training samples for MNIST, 6000 for Fashion-MNIST and 10000 for CIFAR10. The validation is done using the full test dataset (10000 samples for each task).}
  \label{sample-table}
  \centering
  \begin{tabular}{lccccccccc}
    \toprule
     & \multicolumn{3}{c}{MNIST ($r=100$)} & \multicolumn{3}{c}{Fashion-MNIST ($r=20$)} & \multicolumn{3}{c}{CIFAR10 ($r=20$)} \\
     \midrule
    CL & \multicolumn{3}{c}{94.00} & \multicolumn{3}{c}{87.77} & \multicolumn{3}{c}{66.15} \\
    \midrule
     & \scalebox{0.97}{$N=2$} & \scalebox{0.97}{$N=5$} & \scalebox{0.97}{$N=10$} & \scalebox{0.97}{$N=2$} & \scalebox{0.97}{$N=5$} & \scalebox{0.97}{$N=10$} & \scalebox{0.97}{$N=2$} & \scalebox{0.97}{$N=5$} & \scalebox{0.97}{$N=10$} \\
    \midrule
    FL   & 92.64 & 86.79 & 70.06 & 89.79 & 89.28 & 88.21 & 67.99 & 59.18 & 51.05  \\
    \midrule
    IL   & 91.46 & 85.26 & 72.86 & 86.04 & 83.61 & 80.52 & 59.85 & 46.46 & 38.51  \\
    FD   & 94.45 & 90.55 & 77.90 & 87.17 & 83.32 & 79.44 & 56.75 & 44.91 & 31.43  \\
    Ours & 94.19 & 90.63 & 82.07 & 87.91 & 84.44 & 80.77 & 63.49 & 47.28 & 37.78  \\
    \bottomrule
  \end{tabular}
  \label{tab:results}
\end{table}
\paragraph{Effect of $\mathcal{L}_{KD}$}
The feature-based KD term $\mathcal{L}_{KD}$ makes sure that similar feature representations are learned by each client. Fig. \ref{fig:tSNE} provides insight into the evolution of the latent feature space during training. In FL, the structure of the latent feature space is imposed from the start to each client, whereas in our framework, clients have more freedom to learn the best structure that fits their data, since they are only \emph{guided} towards a global representation structure. This clearly improves convergence, as we can see that clusters are already formed at round 25 with our algorithm, whereas with FL the latent feature space is still relatively chaotic at that point. On the other hand, we observe that for IL, the learned latent feature space is totally different for each client as they learn it by themselves, and the convergence rate is also notably slower (compare round 25 for instance).
\paragraph{Effect of $\mathcal{L}_{disc}$}
The effect of the contrastive objective is more subtle, but can still improve the test accuracy significantly if tuned properly. In fact, whereas $\mathcal{L}_{KD}$ can be used without $\mathcal{L}_{disc}$, the reciprocal is not true as observed on Fig. \ref{fig:lambda_ablation}. This is because we have defined the discriminator of user $u$ (i.e., $\hat{h}_u$) using its feature representation classifier $\tau_u$ (see Eq. \eqref{eq:contrastive_classifier}). Hence, in order to generate useful gradients for back propagation (i.e., that do not look like noise), $\tau_u$ must be able to classify relatively well the feature representations that are generated by other models. We have also tested external discriminators (like in \cite{Tian19}), but the results were not as positive.
\paragraph{Limitations and Future work}
As shown by the results, our approach is particularly relevant when the model capacity is relatively small, but struggles when the model complexity grows, especially when compared with FL.
However, the comparison is unfair in that scenario, as the amount of information that is shared between clients at each round is larger by several orders of magnitudes for FL. We also posit that the convergence gap could be reduced by adding sophistication into the algorithm (i.g., by having a local algorithm that selects the most relevant observed representations from other users, by only sharing a fraction of the feature representation dimensions, by introducing a more complex discriminator $h$, etc.). 
Another important limitation is the fact that our framework cannot be used for supervised tasks other than classification, as the feature representations need to be sorted by class. This is harder to alleviate since the whole system's architecture is build upon this requirement. Future work should also investigate how the number of class and their distributions might influence the performance of the algorithm, and also if (and how) it could be used for unsupervised learning.

\section{Conclusion}
\label{sec:conclusion}
We introduce a new collaborative learning algorithm that enables tunable collaboration in cross-device applications and whose uplink and downlink communication does not scale with the model size (as in FL) or the dataset size (as in SL). We prove that our objective is well posed from the point of view of collaboration, as it maximizes a lower bound on the mutual information between the feature representations of different users across the network.  Then, we show empirically that it is particularly relevant in setups where the number of clients is large and when each of them have limited computational resources. In that scenario, it converges faster than IL, FD and even standard FL. 
Due to its simplicity and modularity with other potential requirements (such as differential privacy, server free/peer-to-peer configurations, heterogeneous model architectures, etc.), we posit that our algorithm could be implemented in numerous settings where FL (or other similar frameworks) is not suitable, as long as careful task-specific deployment considerations are made.

\begin{ack}
Living and travel expenses of the main author partially supported by the Hasler Foundation and EPFL.
\end{ack}

\clearpage
{
\small
\bibliographystyle{plainnat}
\bibliography{references}
}

\clearpage
\section*{Checklist}

The checklist follows the references.  Please
read the checklist guidelines carefully for information on how to answer these
questions.  For each question, change the default \answerTODO{} to \answerYes{},
\answerNo{}, or \answerNA{}.  You are strongly encouraged to include a {\bf
justification to your answer}, either by referencing the appropriate section of
your paper or providing a brief inline description.  For example:
\begin{itemize}
  \item Did you include the license to the code and datasets? \answerYes{See Section~\ref{gen_inst}.}
  \item Did you include the license to the code and datasets? \answerNo{The code and the data are proprietary.}
  \item Did you include the license to the code and datasets? \answerNA{}
\end{itemize}
Please do not modify the questions and only use the provided macros for your
answers.  Note that the Checklist section does not count towards the page
limit.  In your paper, please delete this instructions block and only keep the
Checklist section heading above along with the questions/answers below.

\begin{enumerate}

\item For all authors...
\begin{enumerate}
  \item Do the main claims made in the abstract and introduction accurately reflect the paper's contributions and scope?
    \answerYes{}
  \item Did you describe the limitations of your work?
    \answerYes{} See Section \ref{sec:discussion}.
  \item Did you discuss any potential negative societal impacts of your work?
    \answerNA{} 
  \item Have you read the ethics review guidelines and ensured that your paper conforms to them?
    \answerYes{}
\end{enumerate}

\item If you are including theoretical results...
\begin{enumerate}
  \item Did you state the full set of assumptions of all theoretical results?
    \answerYes{} 
        \item Did you include complete proofs of all theoretical results?
    \answerYes{} See Appendix~\ref{sec:proof}.
\end{enumerate}

\item If you ran experiments...
\begin{enumerate}
  \item Did you include the code, data, and instructions needed to reproduce the main experimental results (either in the supplemental material or as a URL)?
    \answerYes{} 
  \item Did you specify all the training details (e.g., data splits, hyperparameters, how they were chosen)?
    \answerYes{} See Section \ref{sec:experiments}.
        \item Did you report error bars (e.g., with respect to the random seed after running experiments multiple times)?
    \answerYes{} Shaded area represents the std over clients (Fig.~\ref{fig:MNIST})
        \item Did you include the total amount of compute and the type of resources used (e.g., type of GPUs, internal cluster, or cloud provider)?
    \answerYes{} See Section \ref{sec:experiments}.
\end{enumerate}

\item If you are using existing assets (e.g., code, data, models) or curating/releasing new assets...
\begin{enumerate}
  \item If your work uses existing assets, did you cite the creators?
    \answerYes{} (MNIST, Fashion-MNIST and CIFAR10 datasets).
  \item Did you mention the license of the assets?
    \answerYes{}
  \item Did you include any new assets either in the supplemental material or as a URL?
    \answerYes{}
  \item Did you discuss whether and how consent was obtained from people whose data you're using/curating?
    \answerNA{} We only use public datasets.
  \item Did you discuss whether the data you are using/curating contains personally identifiable information or offensive content?
    \answerNA{} We only use public datasets.
\end{enumerate}

\item If you used crowdsourcing or conducted research with human subjects...
\begin{enumerate}
  \item Did you include the full text of instructions given to participants and screenshots, if applicable?
    \answerNA{}
  \item Did you describe any potential participant risks, with links to Institutional Review Board (IRB) approvals, if applicable?
    \answerNA{}
  \item Did you include the estimated hourly wage paid to participants and the total amount spent on participant compensation?
    \answerNA{}
\end{enumerate}
\end{enumerate}

\clearpage
\appendix
\section{Appendix}
\subsection{Notation}
\begin{itemize}
    \item $N$: Number of participating users/clients/peers/data owners.
    \item $d$: Raw input data dimension (e.g., $d=3072$ for $32\times32$ RBG images).
    \item $d'$: Latent feature (or feature representation) space dimensionality (i.e., width of last hidden layer).
    \item $C$: Number of class.
    \item $(\mbf{x}, y)\in \R^d \times \{0,...,C-1\}$: Labeled data sample.
    \item $p_{\mbf{X},Y,U}: \mathcal \R^d \times \{1,...,C\} \times \{1,...,U\}\to \R_+$: Joint probability density function across users.
    \item $p_u(\mbf{x}, y) := p_{\mbf{X},Y|U=u}(\mbf{x},y)$: Data distribution of user $u$.
    \item $\mathcal{D}_u := \{(\mbf{x}_i, y_i) \overset{iid}{\sim} p_u\}_{i=1}^{n_i}$: Dataset of user $u$.
    \item $\mbf{w}_u := \{\mbs{\theta}_u, \mbs{W}_u, \mbf{b}_u\}$ with $\mbs{\theta}_u \in \Theta_u, \mbs{W}_u\in \R^{C\times d'}, \mbf{b}_u \in \R^C$: Parameters of the neural network of $u$, with $\Theta_u$ the achievable model parameters for user $u$.
    \item $\phi_u: \R^d \to \R^{d'}, \ \mbf{x} \mapsto \mbf{x}' = \phi(\mbf{x}; \mbs{\theta}_u)$: Neural network of $u$ (or similar parameterized function) that maps a raw input into a latent feature space.
    \item $\tau_u: \R^{d'} \to \R^C, \ \mbf{x}'\mapsto \mbf{z} = \tau(\mbf{x}'; \mbf{W}_u, \mbf{b}_u) := \mbf{W}_u\mbf{x}' + \mbf{b}_u$: Linear classifier of $u$.
    \item $\lambda_{KD}, \lambda_{disc}, n_{avg}, M_\uparrow, M_\downarrow$: Hyperparameters.
    \item $\eta$: Learning rate.
    \item $\ell_{CE}, \ell_{KD}, \ell_{disc}$: Cross-entropy, feature-based KD and discriminator loss functions, respectively.
    \item $\mathcal{L}_{CE}, \mathcal{L}_{KD}, \mathcal{L}_{disc}$: Expected value (over the data) of $\ell_{CE}, \ell_{KD}$ and $\ell_{disc}$, respectively.
    \item $L_{CE}, L_{KD}, L_{disc}$: Mini-batch estimates of $\mathcal{L}_{CE}, \mathcal{L}_{KD}$ and $\mathcal{L}_{disc}$, respectively.
    \item $\mbs{\Phi}_s, \mbs{\Phi}_t$: Random vectors (feature representations) of the student (user $u$) and the teacher (random user $U$).
    \item $p_{s,t}, p_s, p_t$: Joint and marginal distributions of $\mbs{\Phi}_s$ and $\mbs{\Phi}_t$, respectively.
    \item $q$: Joint distribution of $\mbs{\Phi}_s, \mbs{\Phi}_t$ and $I$, where $I$ is a binary random variable indicating if $\mbs{\Phi}_s, \mbs{\Phi}_t$ has been drawn from $p_{s,t}$ ($I=1$) or $p_sp_t$ ($I=0$).
    \item $h$: Binary discriminator with Bernoulli parameter $\hat{h}$ (i.e., learnable estimate of $q_{I|\mbs{\Phi}_s, \mbs{\Phi}_t}$).
    \item $\mbf{s}, \mbf{t}$: Observation/realization of $\mbs{\Phi}_s$ and $\mbs{\Phi}_t$, respectively.
    \item $\bar{\mbf{t}}^c, \mbf{t}^c$: Global (i.e., using all the samples across users) and local (i.e., using $n_{avg}$ samples) average feature representations of class $c$, respectively.
\end{itemize}
\newpage
\subsection{Proof of Theorem \ref{thm:I_bound}}
\label{sec:proof}
Recall that $q(\mbf{s}, \mbf{t}, i)$ is the joint distribution of $(\mbs{\Phi}_s, \mbs{\Phi}_t, I)$ such that  $q(\mbf{s}, \mbf{t}|i=1) = p_{s,t}(\mbf{s}, \mbf{t})$ and $q(\mbf{s}, \mbf{t}|i=0) = p_s(\mbf{s})p_t(\mbf{t})$ and suppose that the prior $q(i)$ satisfy $q(i=1) = \frac{1}{K+1}$ and $q(i=0) = \frac{K}{K+1}$, i.e., for each sample from the distribution $p_{s,t}$, we draw $K$ samples from the distribution $p_s p_t$. We have:
\begin{align}
    \mathcal{I}(\mbs{\Phi}_s, \mbs{\Phi}_t) 
    & =
    \E_{(\mbs{\Phi}_s, \mbs{\Phi}_t)\sim p_{s,t}}\left[-\log\frac{p_s(\mbs{\Phi}_s)p_t(\mbs{\Phi}_t)}{p_{s,t}(\mbs{\Phi}_s, \mbs{\Phi}_t)}\right]
    \\ & = 
    \E_{(\mbs{\Phi}_s, \mbs{\Phi}_t)\sim p_{s,t}}\left[-\log\left(K \frac{p_s(\mbs{\Phi}_s)p_t(\mbs{\Phi}_t)}{p_{s,t}(\mbs{\Phi}_s, \mbs{\Phi}_t)}\right)\right] + \log(K) 
    \\ & \geq 
    \E_{(\mbs{\Phi}_s, \mbs{\Phi}_t)\sim p_{s,t}}\left[-\log\left(1+K \frac{p_s(\mbs{\Phi}_s)p_t(\mbs{\Phi}_t)}{p_{s,t}(\mbs{\Phi}_s, \mbs{\Phi}_t)}\right)\right] + \log(K) 
    \\ & = 
    \E_{(\mbs{\Phi}_s, \mbs{\Phi}_t)\sim p_{s,t}}\left[\log q(i=1 | \mbs{\Phi}_s, \mbs{\Phi}_t)\right] + \log(K) \label{eq:lower_bound_I_1}
\end{align}
where the last equality is obtained using the Bayes' rule on the posterior $q(i=1 | \mbs{\Phi}_s, \mbs{\Phi}_t)$:
\begin{align}
    q(i=1 | \mbs{\Phi}_s, \mbs{\Phi}_t) 
    & = 
    \frac{q(\mbs{\Phi}_s, \mbs{\Phi}_t|i=1)q(i=1)}{q(\mbs{\Phi}_s, \mbs{\Phi}_t|i=0)q(i=0) + q(\mbs{\Phi}_s, \mbs{\Phi}_t|i=1)q(i=1)} 
    \\ & = \frac{p_{s,t}(\mbs{\Phi}_s, \mbs{\Phi}_t)}{Kp_s(\mbs{\Phi}_s)p_t(\mbs{\Phi}_t) + p_{s,t}(\mbs{\Phi}_s, \mbs{\Phi}_t)} 
    \\ & = \left(1+K \frac{p_s(\mbs{\Phi}_s)p_t(\mbs{\Phi}_t)}{p_{s,t}(\mbs{\Phi}_s, \mbs{\Phi}_t)}\right)^{-1}. 
\end{align}
Hence, by optimizing $\E_{(\mbs{\Phi}_s, \mbs{\Phi}_t)\sim p_{s,t}}\left[\log q(i=1 | \mbs{\Phi}_s, \mbs{\Phi}_t)\right]$ with respect to the model parameters $\mbs{\theta}_u$ of the student, we optimize a lower bound on the mutual information between $\mbs{\Phi}_s, \mbs{\Phi}_t$. By noting that $\log q(i=0|\mbs{\Phi}_s, \mbs{\Phi}_s) \leq 0$, we can further bound the expectation term in \eqref{eq:lower_bound_I_1} as follows:
\begin{align}
    \E_{(\mbs{\Phi}_s, \mbs{\Phi}_t)\sim p_{s,t}}\left[\log q(i=1 | \mbs{\Phi}_s, \mbs{\Phi}_t)\right] 
    & \geq 
    \E_{(\mbs{\Phi}_s, \mbs{\Phi}_t)\sim q | I=1}\left[\log q(i=1 | \mbs{\Phi}_s, \mbs{\Phi}_t)\right] \nonumber
    \\ & \quad + 
    K \E_{(\mbs{\Phi}_s, \mbs{\Phi}_t)\sim q | I=0}\left[\log q(i=0 | \mbs{\Phi}_s, \mbs{\Phi}_t)\right]
    \\ & =
    (K+1) \sum_{i}q(i) \E_{(\mbs{\Phi}_s, \mbs{\Phi}_t)\sim q | I=i}\left[\log q(i | \mbs{\Phi}_s, \mbs{\Phi}_t)\right] 
    \\ & =
    (K+1) \E_{(\mbs{\Phi}_s, \mbs{\Phi}_t, I)\sim q }\left[\log q(I| \mbs{\Phi}_s, \mbs{\Phi}_t)\right] 
    \\ & =
    (K+1) \E_{(\mbs{\Phi}_s, \mbs{\Phi}_t)\sim q}\left[\E_{I\sim q|\mbs{\Phi}_s, \mbs{\Phi}_t}\left[\log q(I| \mbs{\Phi}_s, \mbs{\Phi}_t)\right]\right] \label{eq:L_disc} 
\end{align}
However, similar to \citet{Tian19}, the Bernoulli distribution $q(i | \mbs{\Phi}_s, \mbs{\Phi}_t)$ is unknown and must therefore be approximated by training a discriminator $h: \{0,1\} \times \R^{d'} \times \R^{d'} \to [0,1]$. Using Gibbs' inequality, we obtain
\begin{equation}
    -\E_{I\sim q|\mbs{\Phi}_s, \mbs{\Phi}_t}\left[\log q(I| \mbs{\Phi}_s, \mbs{\Phi}_t)\right]
    \leq
    -\E_{I\sim q|\mbs{\Phi}_s, \mbs{\Phi}_t}\left[\log h(I, \mbs{\Phi}_s, \mbs{\Phi}_t)\right],
\end{equation}
where the right-hand term is the expected negative log-likelihood loss of the discriminator for a particular set $(\mbs{\Phi}_s, \mbs{\Phi}_t)$. Hence, Eq. \eqref{eq:L_disc} is proportional to minus the expected loss of the discriminator. Let $\hat{h}(\mbf{s}, \mbf{t})\in [0,1]$ denote the Bernoulli parameter of $h$ given the data $(\mbf{s}, \mbf{t})$ (i.e., ${h(i,\mbf{s}, \mbf{t}) = \hat{h}(\mbf{s}, \mbf{t})^i(1-\hat{h}(\mbf{s}, \mbf{t}))^{1-i}}$),  we define our learning objective for the discriminator as follows:
\begin{align}
-\mathcal{L}_{disc}(h, \phi_u) 
& := 
(K+1)\E_{(\mbs{\Phi}_s, \mbs{\Phi}_t)\sim q}\left[\E_{I\sim q|\mbs{\Phi}_s, \mbs{\Phi}_t}\left[\log h(I, \mbs{\Phi}_s, \mbs{\Phi}_t)\right]\right]
\\ & = 
(K+1) \E_{(\mbs{\Phi}_s, \mbs{\Phi}_t, I)\sim q }\left[\log h(I| \mbs{\Phi}_s, \mbs{\Phi}_t)\right] 
\\ & = 
(K+1)\E_{(\mbs{\Phi}_s, \mbs{\Phi}_t, I)\sim q}\left[\log\left( \hat{h}(\mbs{\Phi}_s, \mbs{\Phi}_t)^I(1-\hat{h}(\mbs{\Phi}_s, \mbs{\Phi}_t))^{(1-I)}\right)\right] 
\\ & = 
+ (K+1)\E_{(\mbs{\Phi}_s, \mbs{\Phi}_t)\sim q|I=1}\left[\log \hat{h}(\mbs{\Phi}_s, \mbs{\Phi}_t)\right]q(i=1) 
\nonumber \\ & \quad 
(K+1)\E_{(\mbs{\Phi}_s, \mbs{\Phi}_t)\sim q | I=0}\left[\log\left(1-\hat{h}(\mbs{\Phi}_s, \mbs{\Phi}_t)\right)\right]q(i=0)
\\ & = 
\E_{(\mbs{\Phi}_s, \mbs{\Phi}_t)\sim p_{s,t}}\left[\log\hat{h}(\mbs{\Phi}_s, \mbs{\Phi}_t)\right]
+ K\E_{(\mbs{\Phi}_s, \mbs{\Phi}_t)\sim p_sp_t}\left[\log(1-\hat{h}(\mbs{\Phi}_s, \mbs{\Phi}_t))\right],
\end{align}
which concludes the derivation.
\subsection{Algorithms}
\begin{algorithm2e}[H]
\label{alg:FRED}
\caption{\textsc{GlobalUpdate}}
\KwIn{Server $S$, $N$ users with local datasets $\{\mathcal{D}_u\}_{u=1}^N$.}
\BlankLine
$S$ initializes randomly $\{\bar{\mbf{t}}^c\}_{c=1}^C$ and random observations $\{\{\mbf{t}_m^c\}_{c=1}^C\}_{m=1}^{N\cdot M_\uparrow}$\;
Each client $u$ initializes $\mbf{w}_u^0:=\{\mbs{\theta}^0_u, \mbf{W}^0_u, \mbf{b}^0_u\}$\;
$r\gets 0$\;
\While{training:}{
$r\gets r+1$\;
\For{each client $u$:}{
$u$ downloads $\{\bar{\mbf{t}}^c\}_{c=1}^C$ and $M_\downarrow$ random observations $\{\{\mbf{t}_m^c\}_{c=1}^C\}_{m=1}^{M_\downarrow}$\;
$\mbf{w}_u^r \gets \mbox{\textsc{LocalUpdate}}(\mathcal{D}_u, \mbf{w}_u^{r-1}, \{\bar{\mbf{t}}^c\}_{c=1}^C, \{\{\mbf{t}_m^c\}_{c=1}^C\}_{m=1}^{M_\downarrow}$)\;
$u$ computes and uploads its local averaged representations $\{\bar{\mbf{t}}_u^c\}_{c=1}^C$\;
$u$ computes (using $n_{avg}$) and uploads $M_\uparrow$ observations $\{\{\mbf{t}_{u,m}^c\}_{c=1}^C\}_{m=1}^{M_\downarrow}$\;}
$S$ aggregates $\{\{\bar{\mbf{t}}_u^c\}_{c=1}^C\}_{u=1}^N$ to obtain $\{\bar{\mbf{t}}^c\}_{c=1}^C$\;
$S$ stores (and shuffles) $\{\{\mbf{t}_{u,m}^c\}_{c=1}^C\}_{m=1}^{M_\downarrow}$ in their corresponding class buffers\;
}
\Return $\{\mbf{w}_u^{r}\}_{u=1}^N$
\end{algorithm2e}
\begin{algorithm2e}[H]
\label{alg:LocalUpdate}
\caption{\textsc{LocalUpdate}}
\KwIn{Local dataset $\mathcal{D}_u$, model parameters $\mbf{w}_u = \{\mbs{\theta}_u, \mbf{W}_u, \mbf{b}_u\}$, global features $\{\bar{\mbf{t}}^c\}_{c=1}^C$ and observations $\{\{\mbf{t}_m^c\}_{c=1}^C\}_{m=1}^{M_\downarrow}$, number of local training rounds $E$, averaging parameter $n_avg $.}
\BlankLine
Buffer initialization (per class): $\{\mbs{\hat{\Phi}}_u^{c} \gets \mbf{0}\}_{c=1}^C$\;
\For{$e \in [1,...,E]$:}{
    \For{mini-batch $\mathcal{B} \in \mathcal{D}_u$:}{
        $L_{KD}, L_{CE}, L_{disc} \gets 0$\;
        \For{$(\mbf{x}_i, y_i) \in \mathcal{B}$:}{
        $\mbf{s}_i \gets \phi_u(\mbf{x}_i)$\;
        $L_{CE} \gets L_{CE} + \frac{1}{|\mathcal{B}|}\ell_{CE}(\tau_u(\mbf{s}_i), y_i)$\;
        $L_{KD} \gets L_{KD} + \frac{1}{|\mathcal{B}|}\ell_{KD}(\mbf{s}_i, \bar{\mbf{t}}^{y_i})$\;
        Sample $m\sim \mbox{\textsc{Uniform}}(1,...,M_\downarrow)$\;
        $L_{disc} \gets L_{disc} + \frac{1}{|\mathcal{B}|}\left(
        \ell_{disc}(1, \mbf{s}_i, \mbf{t}_m^{y_i}) 
        + \sum_{c\neq y_i} \ell_{disc}(0, \mbf{s}_i, \mbf{t}_m^{c}) 
        \right)$\;}
    $L \gets  L_{CE} + \lambda_{KD} L_{KD} + \lambda_{disc} L_{disc}$\;
    $\mbf{w}_u \gets \mbf{w}_u - \eta\nabla_{\mbf{w}_u}L$\;
    }
}
\Return $\mbf{w}_u$
\end{algorithm2e}

\end{document}